\documentclass{article} 
\usepackage{nips14submit_e,times}
\usepackage{url}
\usepackage{amsmath}
\usepackage{bbold}
\usepackage{graphicx}
\usepackage{wrapfig}
\usepackage{listings}
\usepackage[ruled,vlined]{algorithm2e}
\usepackage{booktabs}
\usepackage{tabularx}

\title{GPU based Path Integral Control with Learned Dynamics}

\author{
Grady Williams \\
Autonomous Control and \\ Decision  Systems
  Laboratory \\
Georgia Institute of Technology\\
\text{gradyrw@gatech.edu} \\
\And
Eric Rombokas \\
Mechanical Engineering\\
University of Washington \\
\text{eric@rombokas.com} \\
\AND
Tom Daniel \\
Department of Biology\\
University of Washington \\
\text{danielt@uw.edu} \\
}

\nipsfinalcopy 

\newcommand{\rd}{{\mathrm d}}

\newcommand{\rT}{{\mathrm{T}}}

\begin{document}

\maketitle

\begin{abstract}
We present an algorithm which combines recent advances in model based path integral control with machine learning approaches to learning forward dynamics models. We take advantage of the parallel computing power of a GPU to quickly take a massive number of samples from a learned probabilistic dynamics model, which we use to approximate the path integral form of the optimal control. The resulting algorithm runs in a receding-horizon fashion in realtime, and is subject to no restrictive assumptions about costs, constraints, or dynamics. A simple change to the path integral control formulation allows the algorithm to take model uncertainty into account during planning, and we demonstrate its performance on a quadrotor navigation task. In addition to this novel adaptation of path integral control, this is the first time that a receding-horizon implementation of iterative path integral control has been run on a real system.
\end{abstract}

\section{Introduction}
General motion planning is one of the core problems in robotics and there have been recent successes in this field through the use of model-based optimal control. Some recent examples are \cite{kalakrishnan2011stomp} and \cite{tassa2012synthesis} for legged robot locomation, and \cite{mellinger2011minimum} for aggressive control of autonomous quadrotors. Despite these successes, a major drawback with this approach is the reliance on accurate forward or inverse dynamics models, which can be difficult or impossible to develop through physical analysis. A promising method of dealing with this problem is to learn a dynamics model through interaction with the environment instead of relying on an expert provided physics based model. 

An important issue in this approach is that the learned dynamics only accurately describe a small region of the state space, where the training examples lie. Without taking any extra precautions, the controller assumes that the model accurately describes the dynamics for the entire space. This effect, termed model bias in \cite{Deisenroth11pilco}, is potentially catastrophic and must be dealt with for consistent and safe performance.

Various approaches have been succesful in dealing with model bias and have achieved good performance on simple tasks, the most well known of these is perhaps PILCO  \cite{Deisenroth11pilco} which demonstrated remarkable learning performance on a cart-double-pole swing-up task while learning dynamics as a Gaussian Process. Another recent algorithm is due to Kuindersma et al. \cite{ Kuindersma_etal13} who considers risk control for stochastic optimization and demonstrates good performance on stabilization tasks. In contrast to these works, we consider control for tasks which require more general cost descriptions, such as obstacle navigation.

\subsection{Contributions}
We present an algorithm and computational framework for combining model based receding-horizon path integral control with machine learning algorithms that are capable of learning probabilistic forward dynamics models. Specifically we modify the recently developed receding-horizon formulation of Policy Improvement with Path Integrals ($PI^2$-RH) \cite{RH_PI2} to take into account model uncertainty, and then use Locally Weighted Projection Regression (LWPR) \cite{Vijayakumar_locallyweighted} to learn a probabilistic dynamics model of a nano-quadrotor. The new algorithm performs unprecedented massive sampling of the probabilistic dynamics, but runs in real time due to its implementation on a graphics processing unit (GPU). We report the algorithm's performance controlling the nano-quadrotor on an obstacle avoidance navigation task. Additionally this is the first implementation of receding-horizon $PI^2$ on a real machine.

\section{Learning forward models with LWPR}
Many machine learning algorithms are capable of acting as black box predictors for dynamical systems (see \cite{model_learning_for_control} for an overview), however only a subset are able to give probabilistic predictions at a cheap computational cost. LWPR is an algorithm based on a locally weighted incremental version of partial least squares, and under the classic probabilistic interpretation of weighted least squares it's able to give uncertainty estimates along with a mean prediction. The algorithm learns a set of local linear models and equips each model with a center point in the state space and a distance metric. These values are used to compute predictions and uncertainties as a weighted average of all the models, let $y_i(x)$, $\sigma_i(x)$, and $c_i$ be the mean, standard deviation, and center of the $i_{th}$ local model and let $D_i$ be the distance metric. If there are $L$ local models then the LWPR mean $y(x)$ and variance $\sigma(x)^2$ predictions are given by:

\begin{equation}
\small{y(x) = \sum_{i=1}^L w_{i} y_i },  \quad  \small{\sigma(x)^2 = \sum_{j=1}^L    w_{j} ( (y - y_i)^2 + \sigma_k^2 )}\quad  \text{where}  \quad  w_{j} =  \frac{e^{-\frac{1}{2}(x - c_j)^{\rT} D_j(x - c_j)}}{\sum_{i=1}^L  e^{-\frac{1}{2}(x - c_i)^{\rT} D_j(x - c_i)}}
\end{equation}

The key advantage of LWPR is that the computational complexity of updating the local models and making predictions is linear in the dimension of the state space, as opposed to scaling with the number of training inputs like many other algorithms (e.g Gaussian Processes).

\section{Control}

In this section we briefly review Policy Improvement with Path Integrals ($PI^2$)\cite{theodorou2010reinforcement} and its receding horizon implementation ($PI^2$-RH)\cite{RH_PI2}. We then propose a modification to the algorithm which allows it to take into account model uncertainty.

\subsection{Receding-Horizon path integral control}

Consider a stochastic dynamical system of the form $\rd x = f(x,t)  \rd t + G(x)u(t) \rd t + B(x)  \rd w$, this system has non-linear passive dynamics, but is linear in the control and noise terms. We further assume that the noise is control dependent and that $BB^{\rT} = \lambda GR^{-1}G^{\rT}$, for some $\lambda > 0$. Now assume that we're given a cost function of the form $J(x) = \phi(T) + \int_{0}^{T}{(q(x,t) + u(t)^{\rT}Ru(t))\rd t}$, under these assumptions the stochastic Hamilton-Jacobi-Bellman equation for $V(x) = \min_u\mathbb{E}[J(x)]$ can be related to a path integral (See \cite{theodorou2010reinforcement} for the derivation), and the optimal controls take the form:

\begin{equation}
\label{Equation:OptimalControls}
u_{t_i} = \int{P(\tau_i)u_L(\tau_i)\rd \tau_i}  \quad  \text{and} \quad  P(\tau_i) = \frac{\exp(-\frac{1}{\lambda}\tilde{S}(\tau_i))}{\int{\exp(-\frac{1}{\lambda}\tilde{S}(\tau_i))\rd \tau_i}}
\end{equation}

Where the cost-to-go function is: $ \small{ \tilde{S}(\tau_i) = S(\tau_i) + \sum_{i=1}^N (u_t^{\rT}G_{c}^{\rT} H^{-1}G_{c} u_t  \rd t + u_t^{\rT}G_{c}^{\rT} H^{-1}B  \rd w_t^k )} $.

And the portion of the cost-to-go depending on the passive dynamics is: $S(\tau_i) = \phi_{t_N} + \sum_{j=i}^{N-1}q(x,t) \rd t $ with $H = G_{c} R^{-1} G_{c}^{\rT}$ and ${\bf u_L(\tau_i)} = R^{-1}G_{c}^{\rT}(G_{c} R^{-1}G_c^{\rT})^{-1}B_c(x) \rd w$.   In these equations $\tau = \{x_0, x_1, x_2, ... x_{N-1}\}$ denotes a discretized trajectory, and $\tau_i$ is the remaining piece of the trajectory at the i$^{th}$ timestep, N is the number of timesteps in a trajectory (ie. the number of discrete steps between $0$ and $T$), and $\rd t$ is the size of a single step.

In the $PI^2$ algorithm the optimal controls are computed by sampling from the stochastic dynamics in order to approximate (\ref{Equation:OptimalControls}). It's usually necessary to perform this approximation in an iterative fashion: start with an initial guess, sample possible trajectories from the dynamics (termed ``rollouts'') and compute an approximation to (\ref{Equation:OptimalControls}), the resulting approximation will be an improvement over the initial controls but may not be sufficient. If this is the case the process is repeated until the algorithm converges to a solution, and once the solution is computed the control is executed in an open loop fashion. 

The receding-horizon implementation of $PI^2$ modifies this approach by constantly re-planning while simultaneously executing. Instead of iterating to convergence as in traditional $PI^2$ only a small, pre-defined, number of iterations are performed and then only a single timestep of the resulting trajectory is executed. The algorithm then receives state-feedback and warm-starts its optimization for the next timestep with the un-executed portion of the previous timestep's control plan. This modification transforms $PI^2$ into an implicit feedback controller, and allows it to perform difficult navigation tasks with a simulated quadrotor, which the original $PI^2$ algorithm has difficulty with \cite{RH_PI2}.

\subsection{Accounting for model uncertainty}

In order to combat model bias we modify $PI^2$ in order to take into account the uncertainty of dynamics predictions from a learned probabilistic model. The role of the passive dynamics in evaluating rollouts is in the $S(\tau_i)$ portion of the cost-to-go function. In all previous model-based applications of $PI^2$, $S(\tau_i)$ is evaluated based on a known passive dynamics $f(x,u)$ model. In order to evaluate $S(\tau_i)$ under the uncertain dynamics it makes sense to evaluate $\mathbb{E}[S(\tau_i)|F(x,u)]$, the expectation of the cost-to-go with respect to the uncertain dynamics $F(x,u)$, instead of computing a deterministic evaluation of $S(\tau_i)$ based on the mean prediction. This expectation is analytically intractible, but we can compute a crude Monte-Carlo approximation by taking multiple draws from the probability distribution for each rollout. A similar method for evaluating candidate control plans is considered by Bagnell and Schneider in \cite{Bagnell_2001_3791}, however we employ this strategy on a much larger scale. Under this approach a single rollout becomes a set of samples from a conditional probability distribution (termed sub-rollouts), and the cost-to-go of a rollout is evaluated as the average of the sub-rollouts.

\section{System}
We implemented the modified algorithm on a crazyflie \cite{quadrotor} nano-quadrotor. The crazyflie weighs 19 grams and measures 9 centimeters from rotor to rotor. An onboard radio communicates with the ground station and onboard sensors and processors compute euler angles, euler angle rates, and use a PID controller to achieve commanded attitude rate targets. We used a Vicon Motion Capture system to calculate quadrotor position and velocity. The algorithms runs in realtime on the ground station computer which had an Nvidia GTX 780 Ti with 2880 CUDA cores. This computer was able to perform an iteration using only the mean prediction for 1000 50-step rollouts in about 6.5 milliseconds, and for 1000 rollouts with 32 sub-rollouts an iteration runs in about 18 milliseconds.

\subsection{Models}
We used a standard quadrotor model based on rigid body dynamics to run $PI^2$-RH in order to compare it with the LWPR algorithm. Controls took the form of desired attitude rates $(\phi, \theta, \psi)$, and desired total thrust $(\mathcal{F})$ where the desired rates affect the actual rates according to the equation $\dot{r} = 25(r_{desired} - r_{actual})$.

We used flight examples taken from flights with $PI^2$-RH and the analytic model to train 3 LWPR models with inputs $\phi, \theta, \psi, \mathcal{F}$ and with each output being one of the three cartesian accelerations ($\ddot{x}, \ddot{y}, \ddot{z}$). Then, we replaced the rigid-body formulas for those derivatives with the LWPR models to obtain the learned probabilistic dynamics model.The learned models consisted of 15, 15, and 24 locally linear models for $\ddot{x}, \ddot{y}$, and $\ddot{z}$ respectively. In general the learned models were superior to the analytic model, the propagation errors over 1 second of both the analytic and LWPR model for a representative trial run of the quadrotor are given in Table \ref{Table:Errors}.

\begin{table}[h!]
\begin{tabular*}{\textwidth}{@{\extracolsep{\fill}} l | c| c| c| c| c| c}
          & \multicolumn{3}{c}{Analytic Dynamics} & \multicolumn{3}{c}{LWPR Dynamics}\\\hline
    Position/Velocity Component& $x$   & $y$   & $z$   & $x$   & $y$ & $z$   \\\hline
    Position Error (m) & .38 & .63 & .52 & .27 & .37 & .20  \\ \hline
    Velocity Error (m/s) & 1.01 & 1.71 & 1.25 & .78 & 1.05 & .36 \\
\end{tabular*}
\caption{Average error of euler integration simulation after 1 second over a typical quadrotor run.}
\label{Table:Errors}
\end{table}

\section{Experiments}

In our experiments the quadrotor attempted to navigate between three points while avoiding three obstacles placed in the flight arena. The quadrotor performed 4 laps around these points during a trial, and performed 5 trials in total for each parameter setting. We tested the algorithm using the rigid body model and LWPR models with 1 (just the mean prediction), 4, 8, 16, and 32 sub-rollouts. For an instantaneous state cost we used: $q(x) = (x - W_{x})^2 + (y - W_{y})^2 + 10(z - W_{z})^2 + \frac{1}{5}(\phi^2 + \theta^2 + \psi^2) 
+ \frac{1}{10}(\dot{x}^2 + \dot{y}^2 + \dot{z}^2) + 100\sum_{i=1}^3 \exp(-10(d_{x,i}^2 + d_{y,i}^2)) + 10C$. Here $W_i$ denotes the \emph{i}th component of the point that the quadrotor is moving towards, once the quadrotor comes within a quarter meter of the point the goal is changed to another point. The algorithm is not given advance knowledge that this switch will occur. $d_{x,i}$ is the $x$ distance between the quadrotor and the \emph{i}th obstacle, and similarly for $d_{y,i}$. The variable $C$ is an indicator variable determing whether a crash has occured or not. 

The terminal cost was set to 0, and the cost term: $u_t^{\rT}G_{c}^{\rT} H^{-1}G_{c} u_t  \rd t + u_t^{\rT}G_{c}^{\rT} H^{-1}B  \rd w_t^k$ is considered 0 as well. This is because the term $u_t^{\rT}G_{c}^{\rT} H^{-1}G_{c} u_t  \rd t$ is state independent, so the corresponding term in the denominator can be pushed out of the integral and cancels with the term in the numerator. The second term $u_t^{\rT}G_{c}^{\rT} H^{-1}B  \rd w_t^k$ is not $0$, but since control costs are very small it is close to zero and computational effort can be saved by not computing it, without suffering adverse affects.
 
 We ran the algorithm with the maximal settings that allowed real time computation under the given parameters. This meant that the analytic model, and LWPR models with M=1 and M=4 were allowed to perform two iterations of optimization per timestep while all other settings could only perform 1 iteration in real time. Every trial used a time horizon of 1 second and 50 timesteps per second. The parameters for each setting are given in Table \ref{Table:Parameters}.
 
\begin{table}[h!]
\begin{tabular*}{\textwidth}{@{\extracolsep{\fill}} l | c| c| c| c| c| c}
             & RBD Model & M = 1 & M = 4 & M = 8 & M = 16 & M = 32\\ \hline
    Rollouts & 1000           & 1000  & 1000  & 1000  &970     &950    \\ \hline
    Optimizations per Timestep & 2  & 2  & 2  & 1 & 1 & 1    \\\hline      
 \end{tabular*}
 \caption{Number of rollouts and optimization iterations per timestep possible for the sub-rollout setting.}
 \label{Table:Parameters}
\end{table}

\subsection{Results}

The quadrotor was able to consistently complete the navigation task for all of the settings, except for M = 1 when only the mean prediction of LWPR was used. For this setting the quadrotor was overly-aggressive and unable to consistently stay in the flight volume while avoiding obstacles, out of 7 total trials the quadrotor successfully completed the task 4 times, flew into an obstacles once, and flew out of the flight arena twice. 

\begin{figure}[ht]
\centering
\includegraphics[width=\columnwidth, trim=20mm 150mm 20mm 20mm, clip=true]{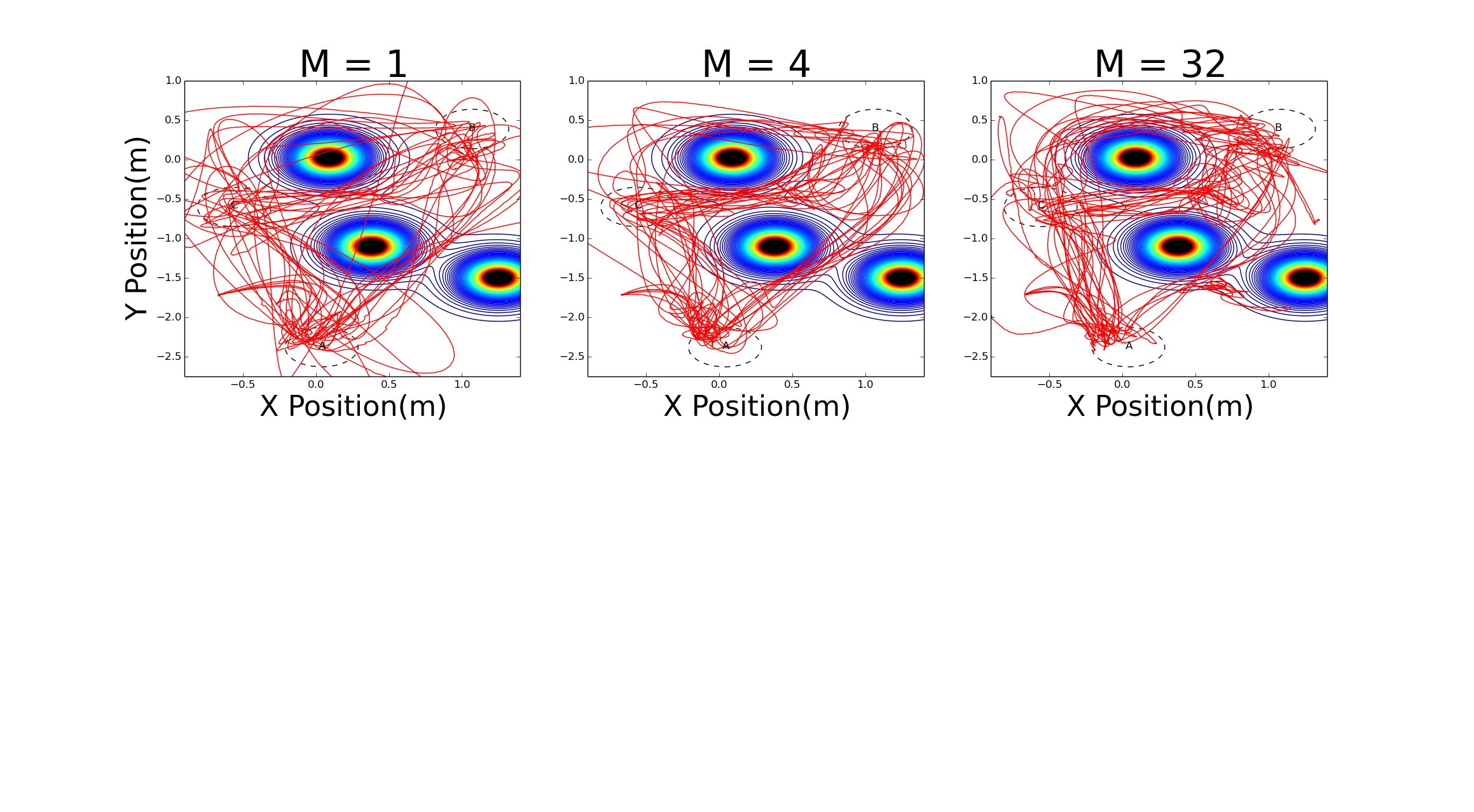}
\caption{Trajectories of all the trials for M = 1 (left), M = 4 (center), and M = 32 (right). The trajectories are overlayed with a contour
         plot of the portion of the cost due to obstacles.}
\label{All of the trial runs for M = 1, M = 4, and M = 32 overlayed on a contour plot of the portion of the cost function due to obstacles.}
\end{figure}

We report the data from the 4 completed trials. Interestingly, when the quadrotor does complete the task for M = 1 the scores are very good. In fact the lowest average for total cost belongs to the succesfully completed trajectories with M = 1. The issue is not poor performance, rather it is overly aggressive and risky performance which comes from not accounting for bias in the learned model.

\begin{table}[h!]
\begin{tabular*}{\textwidth}{@{\extracolsep{\fill}} l | c| c| c| c}
    Sub-Rollouts & Avg. Time & Avg. Total Cost   & Avg. Cost/Sec. & Avg. of 8 Closest Passes (m)\\\hline
    Analytic Model & 37.05  & 6943.27 & 186.93 & .23\\\hline
    M = 1 & 35.39 & 5733.55 & 162.01 & .19 \\ \hline
    M = 4 & 34.91 & 5868.72 & 168.22 & .30 \\\hline
    M = 8 & 47.14 & 7273.80 & 154.27 & .29 \\\hline
    M = 16 & 49.40 & 7491.60 & 152.0 &.32 \\\hline
    M = 32 & 52.13 & 8523.40 & 162.99 & .33 \\\hline
 \end{tabular*}
\caption{Chart of performance metrics for all of the parameter settings. Note the the Cost/Sec is the cost over the optimization horizon.}
\label{Table:Performance}
\end{table}

Our method of taking into account the model uncertainty significantly improves the overall performance and consistency of the vehicle, just using a small number of sub-rollouts (M = 4) allows the vehicle to safely complete the task, and outperforms the analytic model in every metric. Using a greater number of sub-rollouts results in the vehicle choosing more deliberate plans, which have a lower cost over the 1 second planning horizon and stay further away from the obstacles.

The time to completion, total cost, cost over the time horizon, and the average distance of the 8 closest passes to obstacles is given in Table \ref{Table:Performance}. We take the average of the 8 closest passes because there are 8 times that the quadrotor has to cross between (or go very far around) the obstacles during a trial.


\subsection{Implicit variance minimization}

The average variance of the LWPR predictions is given in Table \ref{Table:Variances}. There's a clear trend of decreasing variance in the predictions as the number of sub-rollouts increases, this data in conjunction with the more deliberative actions taken by the parameter settings with higher numbers of sub-rollouts is evidence that the algorithm chooses safer, more certain actions when it has more knowledge about the effects of uncertainty on the system.

\begin{table}[h!]
\begin{tabular*}{\textwidth}{@{\extracolsep{\fill}} l | c| c| c| c| c| c}
    Number of Sub-Rollouts& $\ddot{x}$   & $\ddot{y}$ & $\ddot{z}$   \\\hline
    M = 1 & .27 & .39 & 1.39 \\ \hline
    M = 4  & .25 & .38 & 1.32 \\ \hline
    M = 8  & .22 & .34 & 1.18 \\ \hline
    M = 16 & .21 & .33 & 1.17 \\ \hline
    M = 32 & .20 & .32 & 1.14  \\
    \hline
 \end{tabular*}
\caption{Average variance of acceleration predictions.}
\label{Table:Variances}
\end{table}
 
 It's important to note that this variance minimization is happening without any variance term in the cost function. Minimization is happening implicitly because of the dangerous effects of executing uncertain trajectories in an obstacle filled environment. 

\section{Discussion}

We have developed and tested an algorithm which succesfully combines a modified version of $PI^2$-RH, and LWPR. This is the first time that $PI^2$-RH has been implemented on a real system, and also the first time that it has been combined with a learned dynamics model. Our results show that the $PI^2$-RH algorithm succesfully completes a difficult navigation task using an analytic model, and is outperformed by the new algorithm which uses a learned model and takes into account model bias. The naive implementation of the learned model, where just the mean prediction is used, results in un-reliable and dangerous performance. The new algorithm also minimizes the variance of its trajectories when obstacles are present, despite there being no variance term in the cost function. $PI^2$-RH makes no restrictive assumptions about the form of the system dynamics or cost function, so given an accurate model it can theoretically perform a very wide class of tasks. Used in conjunction with model learning the potential exists for autonomous agents to learn complex tasks completely from scratch in a data efficient manner.

\subsubsection*{Acknowledgments}
This research was supported by ONR grant No. N000141010952 for K Morgansen  and T Daniel Komen Endowed Chair (TLD). We would also like to thank the Veterans Affairs (VA), Rehabilitation Research and Development Center of Excellence for Limb Loss Prevention and Prosthetic Engineering who lent us the use of their Motion Analysis Laboratory.

\end{document}